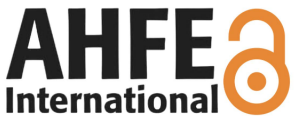

# From Human Narrative to Harmonic Structure: A Human-Centered Investigation of Algorithmic Music Generation through the Chord Wheel Diagram

**Josef Pavlíček[1], Petra Pavlíčková[2] and Irena Štrausová[3]**

[1]Faculty of Information Technology, CTU, Thákurova 9, Prague 6, 160 00, Czech Republic, Josef.Pavlicek@fit.cvut.cz

[2]Faculty of Information Technology, CTU, Thákurova 9, Prague 6, 160 00, Czech Republic, Petra.Pavlickova@fit.cvut.cz

[3]University of South Bohemia in České Budějovice, Faculty of Economics, Department of Applied Mathematics and Informatics, Branišovská 1645/31a, České Budějovice, Czech Republic, istrausova@ef.jcu.cz

## ABSTRACT

Contemporary algorithmic and AI-based music generation can produce compositions that satisfy formal requirements of tonality, harmonic progression, and musical coherence. Nevertheless, structural plausibility raises a fundamental question: whether musical expression can be adequately described by formal and mathematical properties alone. Human composers create music within personal, emotional, and cultural contexts that may influence decisions concerning harmony, repetition, tension, resolution, and deliberate departures from established patterns. This study investigates six narrative-driven popular songs by Bob Dylan, Johnny Cash, and Ritchie Valens with documented compositional, biographical, or cultural contexts. Original human harmonies are compared with outputs of an explainable computational harmonizer operating on the same melodies without access to the original chord progressions. Harmonic vocabulary, functional persistence, repetition, non-diatonic events, and tension–resolution patterns are examined. The Chord Wheel Diagram and BPMN-based representation provide complementary interpretable views: tonal-functional space and temporal harmonic development. Results show that high melody–chord compatibility does not necessarily imply preservation of the original human harmonic decision pattern. Some generated harmonizations retain the economical functional structure of the reference, whereas others increase harmonic diversity, reduce characteristic persistence, or suppress distinctive harmonic events while remaining highly compatible with the melody. Rather than quantifying artistic quality or assuming causal relationships between narrative and individual chords, the study introduces narrative-conditioned harmonic structure as a complementary perspective for computational music analysis. The findings suggest that generative systems may benefit from modeling not only harmonic correctness, but also structural identity, context, and human compositional intention.

## INTRODUCTION

Recent advances in algorithmic and artificial-intelligence-based music generation have substantially increased the ability of computational systems to produce harmonically coherent material. Tonal harmony contains measurable regularities in chord frequency, transition probability, temporal directedness, and hierarchical organization(Moss *et al.*, 2019), while research on harmonic expectation indicates that deviations from statistically expected behavior can contribute to musical perception and preference (Miles, Rosen and Grzywacz, 2017). However, structural plausibility does not necessarily explain why a human composer selects one musically acceptable harmonization rather than another. A melody can support several harmonically valid alternatives. Human harmonic organization therefore involves not only chord–melody compatibility, but also persistence, repetition, harmonic rhythm, tension and resolution, functional transitions, and deliberate departures from expected patterns. The present study describes these recurrent structural characteristics as harmonic decision patterns. From a human-centered perspective, such decisions are made within personal, historical, and cultural contexts. We therefore introduce narrative-conditioned harmonic structure (NCHS): harmonic characteristics that can be examined in relation to documented compositional, biographical, or cultural context without assuming that the context deterministically caused individual harmonic choices.

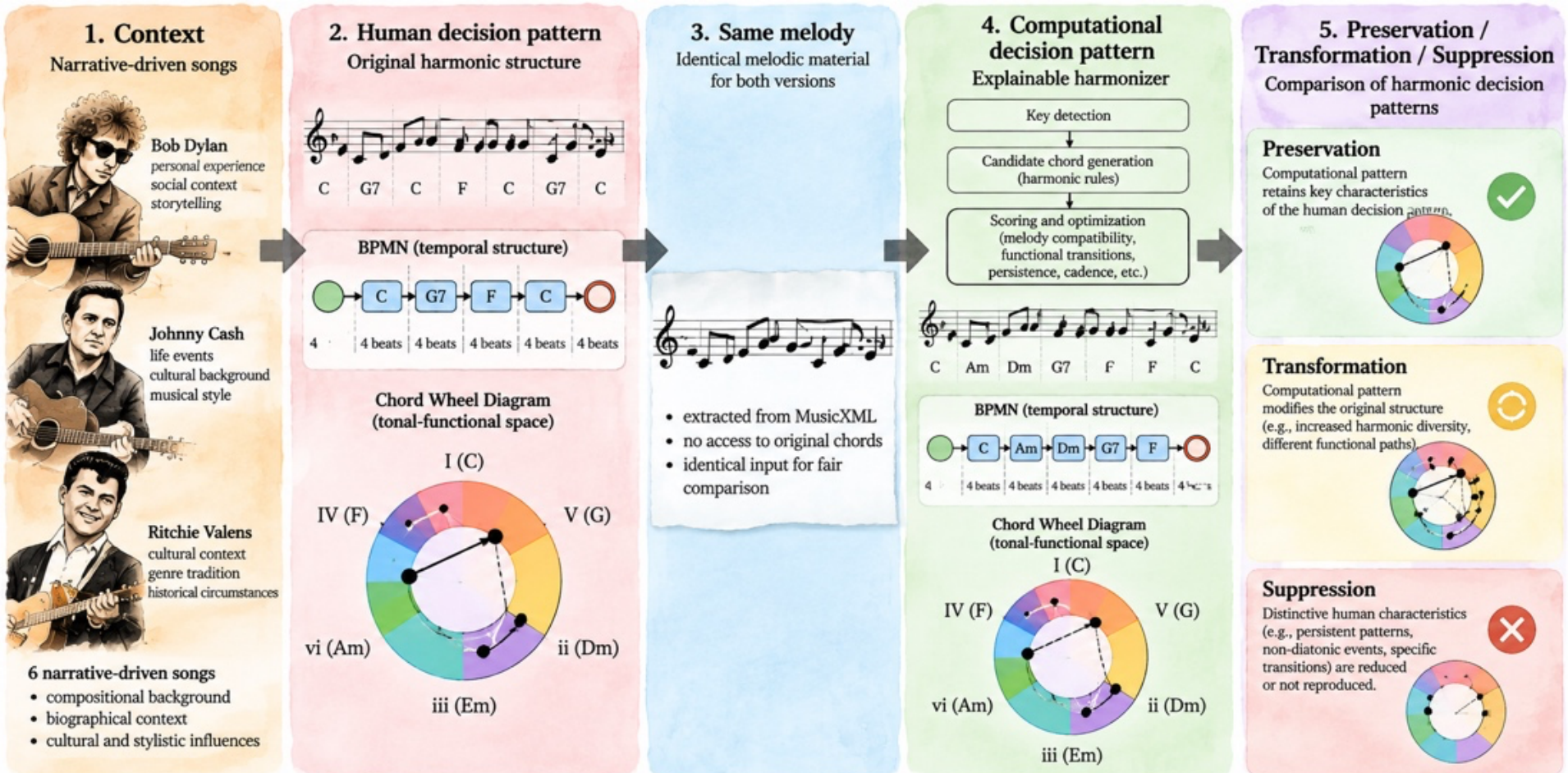

**Figure 1:** Research model linking narrative context, human harmonic decision patterns, shared melodic material, and computational harmonization to the analysis of preservation, transformation, and suppression. The Chord Wheel Diagram is available at www.aakordy.cz. Illustration generated with the assistance of generative AI.

The study compares human and computational harmonic decision-making under controlled melodic conditions. Six narrative-driven popular songs by Bob Dylan, Johnny Cash, and Ritchie Valens are examined. For each composition, the human harmony provides a reference pattern, while the same melody, without the original chords, is processed by an explainable computational harmonizer using explicit criteria for melody–chord compatibility, functional transition, persistence, cadence, and metrical position. The resulting harmonic paths are compared to identify which characteristics of the human reference are preserved, transformed, expanded, or suppressed. The overall research model is summarized in Figure 1.
Interpretability is central to this comparison. Building on previous work by the authors (Pavlicek *et al.*, 2024; Pavlicek, Molhanec and Pavlíčková, 2025; Pavlicek, Pavlickova and Strausova, 2026), the study combines BPMN and the Chord Wheel Diagram as complementary representations. BPMN represents the temporal dimension—when harmonic decisions occur and how long they persist—whereas the Chord Wheel Diagram represents the tonal-functional dimension—where these decisions are situated within harmonic space. The study is guided by the following research questions:

**RQ1:** What identifiable patterns of harmonic decision-making occur in songs with documented narrative or compositional contexts?
**RQ2:** How are these characteristics transformed when the same melodic material is subjected to computational harmonization?
**RQ3:** To what extent does visual representation through the Chord Wheel Diagram and BPMN-based process representation support users in understanding and interpreting harmonic relationships?

The contribution of this work is a human-centered analytical framework that distinguishes harmonic plausibility from preservation of human harmonic decision-making. Rather than evaluating whether human or computational harmony is “better,” the study investigates which structural characteristics make a harmonic decision pattern distinctive and whether those characteristics remain identifiable after computational reharmonization.

# Related Work and Conceptual Background

## Computational Representation of Harmony

Computational approaches to harmony are based on the assumption that at least part of musical organization can be represented through formal relationships between pitches, chords, tonal functions, and their temporal transitions. Corpus-based studies have demonstrated that tonal harmony exhibits measurable regularities in chord occurrence and sequential organization, making harmonic structure suitable for statistical and computational modeling(Moss *et al.*, 2019). Related research has shown that harmonic expectation can be described in terms of probability and surprise, and that these properties are relevant to human perception of musical sequences (Miles, Rosen and Grzywacz, 2017). These principles are also fundamental to algorithmic harmonization. Given a melody, a computational system can identify candidate chords compatible with its notes and select a sequence according to harmonic, probabilistic, or optimization criteria. Such approaches make it possible to generate structurally coherent alternatives; however, the existence of several acceptable harmonizations for the same melody also exposes an important limitation. **Harmonic compatibility describes what** **can work, but does not necessarily explain why a human composer selected one particular solution.**

This distinction motivates the present research. Instead of evaluating generated harmony exclusively according to formal correctness, the study examines whether a computational system reproduces the characteristic *pattern of decisions* observable in the human reference.

## From Harmonic Structure to Harmonic Decision-Making

Human harmonic organization can be characterized not only by the individual chords used, but also by their persistence, recurrence, ordering, functional relationships, harmonic rhythm, and deliberate departures from expected patterns. Musical expectation research indicates that listeners respond to the interaction between expected and unexpected harmonic events rather than simply to conformity with a fixed set of harmonic rules (Miles, Rosen and Grzywacz, 2017). For this reason, the present study uses **harmonic decision pattern** as its principal analytical unit. A harmonic decision pattern describes how a composition organizes harmonic choices over time, including the size of its harmonic vocabulary, tonic–predominant–dominant relationships, repetition and persistence, non-diatonic events, and tension–resolution behavior. A progression based predominantly on three chords may therefore contain a strong and identifiable decision pattern even though its harmonic vocabulary is small. This perspective is extended through the concept of **narrative-conditioned harmonic structure (NCHS)**. NCHS refers to harmonic characteristics that can be examined in relation to documented compositional, biographical, or cultural context. The concept deliberately does not assume a deterministic relationship between narrative and individual harmonic events. Rather, narrative provides a contextual layer within which recurring structural decisions can be interpreted. This distinction is important because the purpose of the analysis is not to infer a composer's psychological state from a chord progression, but to investigate whether identifiable characteristics of a human harmonic solution remain present when the same melody is reharmonized computationally.

## Human-Centered and Explainable Representation of Harmony

The comparison of human and computational decisions also requires an interpretable representation. Previous work by the authors has explored process-oriented visualization of harmonic progressions using BPMN and the representation of tonal-functional relationships through the **Chord Wheel Diagram** (Pavlicek *et al.*, 2024; Pavlicek, Molhanec and Pavlíčková, 2025; Pavlicek, Pavlickova and Strausova, 2026). Subsequent work investigated these representations from a human-centered perspective, including usability and eye-tracking-based evaluation (Pavlíček, Pavlíčková and Štrausová, 2026).The two representations address different dimensions of the same harmonic decision process. **BPMN represents WHEN a harmonic decision occurs**, its duration, sequence, repetition, and temporal relationship to surrounding events. In contrast, the **Chord Wheel Diagram represents WHERE the decision occurs within** **tonal-functional space**, making relationships between tonic, predominant, dominant, and related harmonic alternatives visually accessible. Their combination is particularly relevant to explainable computational generation. A generated chord sequence can be formally correct while differing substantially from the human reference in persistence, harmonic rhythm, functional trajectory, or tonal distance. Representing these differences explicitly makes it possible to move beyond a binary distinction between "correct" and "incorrect" harmonization and instead examine **what has been preserved, what has been transformed, and what has been suppressed**. This provides the conceptual foundation for the experimental comparison presented in the following section.

# Materials and Methods

## Dataset and Case Selection

The study uses six popular songs by three composers representing different forms of documented narrative context. Two compositions were selected for each artist to allow limited within-author comparison while maintaining a sufficiently compact case-oriented dataset. The selected compositions represent personal and relationship narratives, social context, and cultural transformation. The purpose of the selection is not to establish statistical representativeness of popular music, but to provide contrasting cases in which the relationship between documented context and harmonic decision patterns can be examined in detail. The overall experimental procedure follows the research model shown in Figure 1. For each composition, the original human harmony is first represented as a sequence of harmonic decisions. The melody is then separated from this reference harmony and processed by the same explainable computational harmonizer without access to the original chord progression. Human and computational solutions are subsequently compared in terms of functional structure, persistence, repetition, harmonic rhythm, non-diatonic events, and tension–resolution behavior.

**Table 1.** Dataset and contextual dimensions

| Composer | Composition | Contextual dimension | Main harmonic characteristic examined |
|---|---|---|---|
| Bob Dylan | Tangled Up in Blue | relationship, memory, shifting viewpoint | recurrent I–♭VII behavior and non-diatonic movement |
| Bob Dylan | Blowin' in the Wind | social and cultural context | economical I–IV–V functional space |
| Johnny Cash | I Walk the Line | commitment and personal narrative | tonal movement and dominant–tonic organization |
| Johnny Cash | Folsom Prison Blues | confinement, guilt, freedom/mobility | tonic persistence and blues-derived I–IV–V structure |

| Ritchie Valens | Donna | personal relationship | repetition and economical I–IV–V vocabulary |
|---|---|---|---|
| Ritchie Valens | La Bamba | cultural transformation | persistent repetitive I–IV–V loop |

The contextual dimensions in Table 1 are grounded in documented accounts rather than inferred from the harmonic material itself. For example, Cash explicitly described I Walk the Line as both a love song and a personal reminder to "play it straight" (Cash, 1997). Dylan described Tangled Up in Blue in terms of changing temporal and narrative perspectives, while his comments on Blowin' in the Wind explicitly situated its questions in relation to war and social responsibility. Cash described I Walk the Line as both a love song and a personal reminder to "play it straight," whereas Folsom Prison Blues was inspired by his encounter with a film depicting prison life. In Valens' case, Donna was written for Donna Ludwig, while La Bamba represents his rock-and-roll adaptation of a traditional song rooted in the musical culture of Veracruz. These documented contexts are used as interpretive categories rather than as evidence that particular narrative events caused individual harmonic choices.

## Human Harmonic Decision Representation

For each song, the reference harmony is normalized to Roman-numeral functions relative to its tonal center. This allows harmonic patterns to be compared independently of absolute key. The harmonic sequence of composition s is represented as $H_s = \{h_1, h_2, \ldots, h_n\},$ where each harmonic decision $h_i$ contains a chord function, onset position, and duration: $h_i = (c_i, t_i, d_i).$

Here, $c_i$ denotes the normalized harmonic function, $t_i$ its temporal position, and $d_i$ its duration. This representation intentionally preserves repeated harmonic events because persistence itself constitutes part of the decision pattern. For every reference progression, the analysis records harmonic vocabulary, functional transitions, tonic persistence, repetition, harmonic rhythm, non-diatonic events, and cadential behavior. BPMN provides the temporal representation of these decisions, while the Chord Wheel Diagram provides their tonal-functional interpretation.

### Explainable Computational Harmonization

The computational condition is produced by the same explainable harmonization system for all six compositions. The system receives only the monophonic melody extracted from MusicXML and does not have access to the human reference chords. Candidate harmonies are generated for meter-aware harmonic decision segments and evaluated using explicit musical criteria. For a candidate chord c at decision position t, the local evaluation can be expressed as a weighted scoring function: $L_t(c) = w_m M_t(c) + w_p P_t(c) + w_r R_t(c) + w_k K_t(c),$ where $M_t$ represents melody–chord compatibility, $P_t$ harmonic persistence, $R_t$ metrical or rhythmic suitability, and $K_t$ cadential contribution. The weights $w_m, \ldots, w_k$ control the relative influence of the individual criteria. Importantly, the algorithm does not select each chord independently. The generated harmony is treated as a path through a sequence of harmonic decision spaces. For a candidate path $H' = (c_1, c_2, \ldots, c_T)$, the preferred computational solution is

$$H^* = \arg\max_{(H')} \left[ \Sigma_{t=1}^{T} L_t(c_t) + \Sigma_{t=2}^{T} F(c_{t-1}, c_t) \right],$$

where $L_t(c_t)$ represents the local score of chord $c_t$ at decision position t, including melody–chord compatibility, persistence, metrical suitability, and cadential contribution, while $F(c_{t-1}, c_t)$ evaluates the functional relationship between two consecutive harmonic states. The first sum begins at $t = 1$ because every harmonic decision has a local score; the transition term begins at $t = 2$ because the first transition occurs between $c_1$ and $c_2$. A dynamic-programming/Viterbi-like procedure is used to identify the preferred path. This formulation makes the harmonization process reproducible and reflects the fact that the best sequence cannot necessarily be obtained by selecting the locally highest-scoring chord at each position.

**Human–Computational Comparison**

The purpose of the comparison is not to determine whether the computational harmony is better or worse than the original. Instead, the analysis evaluates the extent to which the harmonic decision pattern of the human reference is retained.

For each composition, human harmony $H_s$ and generated harmony $H_s^*$ are compared along five structural dimensions: $D_s = (V_s, P_s, R_s, N_s, T_s)$, where $V_s$ represents harmonic vocabulary, $P_s$ functional persistence, $R_s$ repetition and harmonic rhythm, $N_s$ non-diatonic behavior, and $T_s$ functional transition and tension–resolution structure. Observed differences are subsequently interpreted using the three categories introduced in Figure 1: preservation, when a characteristic human decision pattern remains identifiable in the computational solution; transformation, when the underlying functional principle remains present but its realization changes; and suppression, when a distinctive characteristic of the human reference is reduced or absent from the generated solution. Melody–chord compatibility is reported separately. This distinction is methodologically important because a high compatibility score indicates that the generated harmony fits the melody according to the computational model, but does not establish similarity to the human reference.

## Human-Centered Interpretation

The visual comparison uses the complementary representations developed in our previous research. BPMN represents the temporal dimension of harmonic decision-making, including sequence, duration, repetition, and cadence placement. The Chord Wheel Diagram represents the corresponding tonal-functional relationships. This combined representation supports RQ3 by making differences between human and computational harmonic decisions inspectable without reducing them to a single numerical similarity score. Existing user-study data from the Chord Wheel Diagram research are used as independent evidence concerning the interpretability of the representation; they are not used to establish a preference for either human or computational harmonization.

# Results

The six computational harmonizations differed substantially in the extent to which they retained the human harmonic decision patterns. Table 2 summarizes the detected tonal center, number of decision segments, mean melody–chord compatibility, generated functional vocabulary, and principal relationship to the human reference. Melody–chord compatibility is treated as an internal measure of melodic fit rather than similarity to the original harmony.

***Table 2.*** *Summary of computational harmonization results and principal human–computational differences.*

| Composition | Detected key | Segments | Mean melody fit | Generated functional vocabulary | Main comparison with human reference |
|---|---|---|---|---|---|
| Tangled Up in Blue | A major | 20 | 0.938 | I, ii7, IV, V, vi, vii°, viiø7 | High melodic fit, but recurrent human I–♭VII behavior is suppressed by a more diatonic generated path. |
| Blowin' in the Wind | D major | 64 | 0.878 | I, IV, V | Strong preservation of the economical I–IV–V functional space, with finer reference details simplified. |
| I Walk the Line | E major | 32 | 0.891 | I, ii7, iii, IV, V, V7, vii° | Core tonic–dominant organization remains identifiable, while the generated solution broadens the local functional vocabulary. |
| Folsom Prison Blues | G major | 34 | 0.939 | I, ii7, IV, V, V7, vii° | Very high melodic fit, but reduced tonic persistence and introduction of additional functional alternatives. |
| Donna | F major | 48 | 0.676 | I, ii, iii, IV, V, vi, vii°, viiø7 | Substantial expansion beyond the economical and repetitive human I–IV–V7 reference pattern. |
| La Bamba | F major | 14 | 0.900 | I, ii7, iii, IV, V, vi, viiø7 | Core I–IV–V organization is preserved, but the generated vocabulary expands beyond the highly repetitive human loop. |

Blowin' in the Wind showed the strongest preservation of the predominant I–IV–V organization. La Bamba also retained its core I–IV–V structure but expanded it with additional functions, while Donna showed a stronger expansion beyond the repetitive human reference. The clearest separation between melodic fit and preservation occurred in Folsom Prison Blues and Tangled Up in Blue. Despite high mean compatibility scores of 0.939 and 0.938, respectively, Folsom reduced characteristic tonic persistence, while Tangled Up in Blue largely suppressed the recurrent human I–♭VII relationship. I Walk the Line retained a recognizable tonic–dominant organization while broadening the local functional vocabulary. Across the cases, preservation, transformation, expansion, and suppression could therefore coexist within a single generated solution.

# Discussion

The results distinguish harmonic plausibility from preservation of human harmonic decision-making. Regarding RQ1, human decision patterns involve not only chord vocabulary but also persistence, repetition, functional trajectory, and distinctive non-diatonic events. For RQ2, computational reharmonization may preserve some of these characteristics while transforming or suppressing others. This is particularly evident in Folsom Prison Blues and Tangled Up in Blue, where high melody–chord compatibility coexists with substantial changes to distinctive human decision patterns. NCHS therefore serves as an interpretive rather than causal concept: documented personal, compositional, or cultural context contextualizes structural decisions without attributing individual chords to narrative causes. Repetition and persistence may themselves contribute to structural identity and can be weakened by optimization-oriented harmonization. Regarding RQ3, BPMN and the Chord Wheel Diagram(www.aakordy.cz) provide complementary temporal and tonal-functional views. For music educators, composers, and other domain experts working with generative AI, these visualizations make algorithmic decisions more inspectable by showing what was generated, when it occurs, and where it lies within tonal-functional space. Previous user-centered evaluation supports their interpretability, although the present experiment does not directly test their effect on human–computational comparison. Limitations include the six-song dataset and the harmonizer's candidate vocabulary, scoring criteria, and weights. Future work should test larger corpora and investigate whether structural-identity or contextual descriptors can be incorporated directly into explainable generative models.

# Conclusion

This study compared human harmonic decision patterns with explainable computational harmonizations generated from the same melodies. The results show that a generated harmony can remain melodically and tonally plausible while altering persistence, repetition, functional trajectories, harmonic vocabulary, or distinctive non-diatonic events. Harmonic plausibility therefore does not necessarily imply preservation of human harmonic decision-making. Narrative-Conditioned Harmonic Structure, combined with BPMN and the Chord Wheel

Diagram, provides a human-centered framework for examining which characteristics of a human harmonic solution are preserved, transformed, or suppressed. Future generative systems may benefit from modeling such structural identity alongside harmonic correctness.

# Acknowledgements and Data Availability

Generative AI tools were used during the preparation of this manuscript for language correction and assistance in the preparation of graphical material. The scientific concept, methodology, analysis, interpretation of results, and final manuscript decisions remain the responsibility of the authors. The Explainable Melody Harmonizer developed for this study, together with the source data used to reproduce the computational experiments, is publicly available in the Explainable_Melody_Harmonizer directory of the:
https://github.com/JosefPavlicek/quantum-inspired-music-research.git